\documentclass[11pt]{article}
\usepackage{amsmath}    % align, xrightarrow
\usepackage{amssymb}    % mathbb
\usepackage{booktabs}   % toprule, midrule, bottomrule、
\usepackage{array}
\newcolumntype{C}[1]{>{\centering\arraybackslash}p{#1}}
\usepackage{multirow}
\usepackage[final]{emnlp26}
\usepackage{latexml} % HTML-only front-matter fixes; inert under PDFLaTeX
\usepackage{pifont}

\newcommand{\cmark}{\ding{51}}
\newcommand{\xmark}{\ding{55}}
\usepackage{xcolor}

\usepackage{enumitem}
\usepackage{subcaption}

\usepackage{times}
\usepackage{latexsym}

\usepackage[T1]{fontenc}
\usepackage[utf8]{inputenc}

\usepackage{microtype}

\usepackage{inconsolata}

\usepackage{graphicx}

\title{SlimKV: Joint Token-Feature KV Cache Compression with Reconstruction-Free Beacon Attention}

\iflatexml
\author{
  \textbf{Zihan Teng\textsuperscript{1,2,3}},
  \textbf{Jiayu Zhao\textsuperscript{1,2}},
  \textbf{Wentao Ren\textsuperscript{2}},
  \textbf{Minhao Fan\textsuperscript{2}},
\\
  \textbf{Tianrui Ma\textsuperscript{2}},
  \textbf{Song Chen\textsuperscript{1}\textsuperscript{*}},
  \textbf{Weichen Liu\textsuperscript{2}\textsuperscript{*}}
\\
  \textsuperscript{1}University of Science and Technology of China
\\
  \textsuperscript{2}Nanyang Technological University,
  \textsuperscript{3}Tencent
}
\else
\author{
  \textbf{Zihan Teng\textsuperscript{1,2,3}},
  \textbf{Jiayu Zhao\textsuperscript{1,2}},
  \textbf{Wentao Ren\textsuperscript{2}},
  \textbf{Minhao Fan\textsuperscript{2}},
\\
  \textbf{Tianrui Ma\textsuperscript{2}},
  \textbf{Song Chen\textsuperscript{1,*}},
  \textbf{Weichen Liu\textsuperscript{2,*}}
\\
  \textsuperscript{1}University of Science and Technology of China
\\
  \textsuperscript{2}Nanyang Technological University,
  \textsuperscript{3}Tencent
\\
  \texttt{zihant@mail.ustc.edu.cn, songch@ustc.edu.cn, liu@ntu.edu.sg}
}
\fi

\begin{document}
\maketitle
\iflatexml
\begin{center}
\small
Contact emails: Zihan Teng (\texttt{zihant@mail.ustc.edu.cn}),
Song Chen (\texttt{songch@ustc.edu.cn}), and
Weichen Liu (\texttt{liu@ntu.edu.sg}).\\
\textsuperscript{*}Corresponding authors: Song Chen and Weichen Liu.
\end{center}
\else
\begingroup
\renewcommand{\thefootnote}{*}
\footnotetext{Corresponding authors.}
\endgroup
\fi

\begin{abstract}
Long-context LLM serving is increasingly bottlenecked by KV-cache memory, especially in resource-constrained scenarios. Among existing KV-cache compression strategies, token-wise methods reduce cached states but risk information loss through eviction or condensation, while feature-wise methods reduce per-token KV dimensions but can require full-dimensional reconstruction to apply positional embedding, limiting decoding speedups. We introduce SlimKV, a question-agnostic joint token-feature KV-cache compression method. SlimKV uses low-rank-aware training to compress long contexts into beacon memory states with latent KV representations, together with layer-adaptive rank allocation. We further uncover a positional asymmetry: removing key-side RoPE affects beacon and raw tokens differently, with much smaller degradation for beacon tokens. Exploiting this asymmetry, SlimKV trains beacon KV projections under a K-RoPE-free constraint and enables latent-space attention during decoding, mitigating reconstruction latency. On LongBench, SlimKV outperforms baselines at 16$\times$/32$\times$ compression and remains leading at 4$\times$/8$\times$, where it retains over 96\% of the uncompressed model's score. Needle-in-a-Haystack confirms robustness across evidence positions, and efficiency evaluation shows up to 7.34$\times$ attention speedup and 3.38$\times$ end-to-end decoding speedup over the uncompressed model at 128K length.
\end{abstract}

\section{Introduction}
\begin{figure}[!t]
    \centering
    \includegraphics[width=1\linewidth]{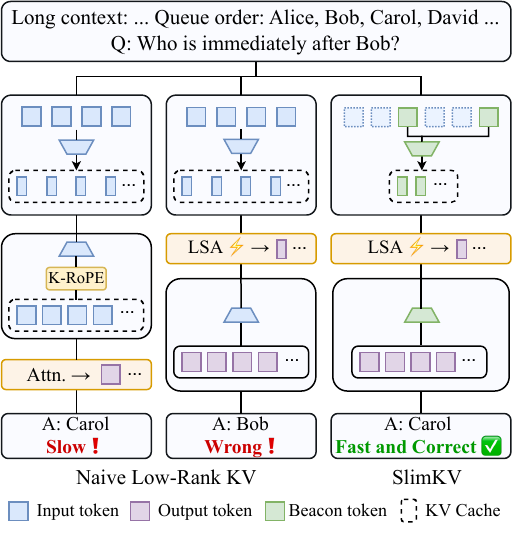}
    \caption{Comparison of low-rank KV-cache strategies. K-RoPE forces reconstruction, while its removal enables latent-space attention (LSA) but loses order information. SlimKV uses K-RoPE-free beacon tokens to retain semantics and positional cues in latent KV space.}
    \label{fig:placeholder}
\end{figure}

Large language models (LLMs) have shown strong performance across diverse tasks, and long-context capability is crucial for document understanding, multi-turn interaction, and complex reasoning~\citep{tworkowski2023focused,liu2025comprehensive}. However, during long-context inference, the key-value (KV) cache grows rapidly with context length, causing substantial memory pressure and memory-access latency \citep{xiao2024streamingllm}. KV-cache overhead therefore becomes a major challenge for efficient deployment under limited memory and bandwidth~\citep{kwon2023efficient}.

Two common directions for long-context KV-cache compression are token-wise and feature-wise. Token-wise methods reduce cached state counts, but explicit deletion can cause substantial information loss under aggressive compression \citep{kim2025kvzip}, and eviction-based methods often require full KV construction before deletion, limiting prefill-compute savings \citep{zhang-2023-h2o,li2024snapkv}. Feature-wise methods reduce cached state dimensions, but the compressed Keys often require full-dimensional reconstruction before applying RoPE, introducing a latency bottleneck \citep{saxena-etal-2024-eigen,chang2025palu}. Activation Beacon \citep{zhang-2025-ActivationBeacon} alleviates severe information loss under aggressive compression by softly compressing historical context into beacon states. However, the redundancy of these beacon states has not been further explored. More importantly, the Key-reconstruction latency bottleneck of feature-wise compression remains unresolved. This motivates our central question:

\vspace{0.5em}
\emph{Can joint token-feature compression overcome single-axis limits while avoiding the reconstruction bottleneck?}
\vspace{0.5em}

To answer this question, we propose \textbf{SlimKV}, a joint token-feature KV-cache compression framework built upon beacon-based soft compression~\citep{zhang-2025-ActivationBeacon}. SlimKV trains beacon KV projections directly under low-rank constraints, allowing each beacon state to summarize historical tokens while forming compact latent KV representations. To better utilize a fixed feature budget, SlimKV further allocates adaptive ranks according to layer-wise low-rank heterogeneity.

\begin{figure}
    \centering
    \includegraphics[width=1\linewidth]{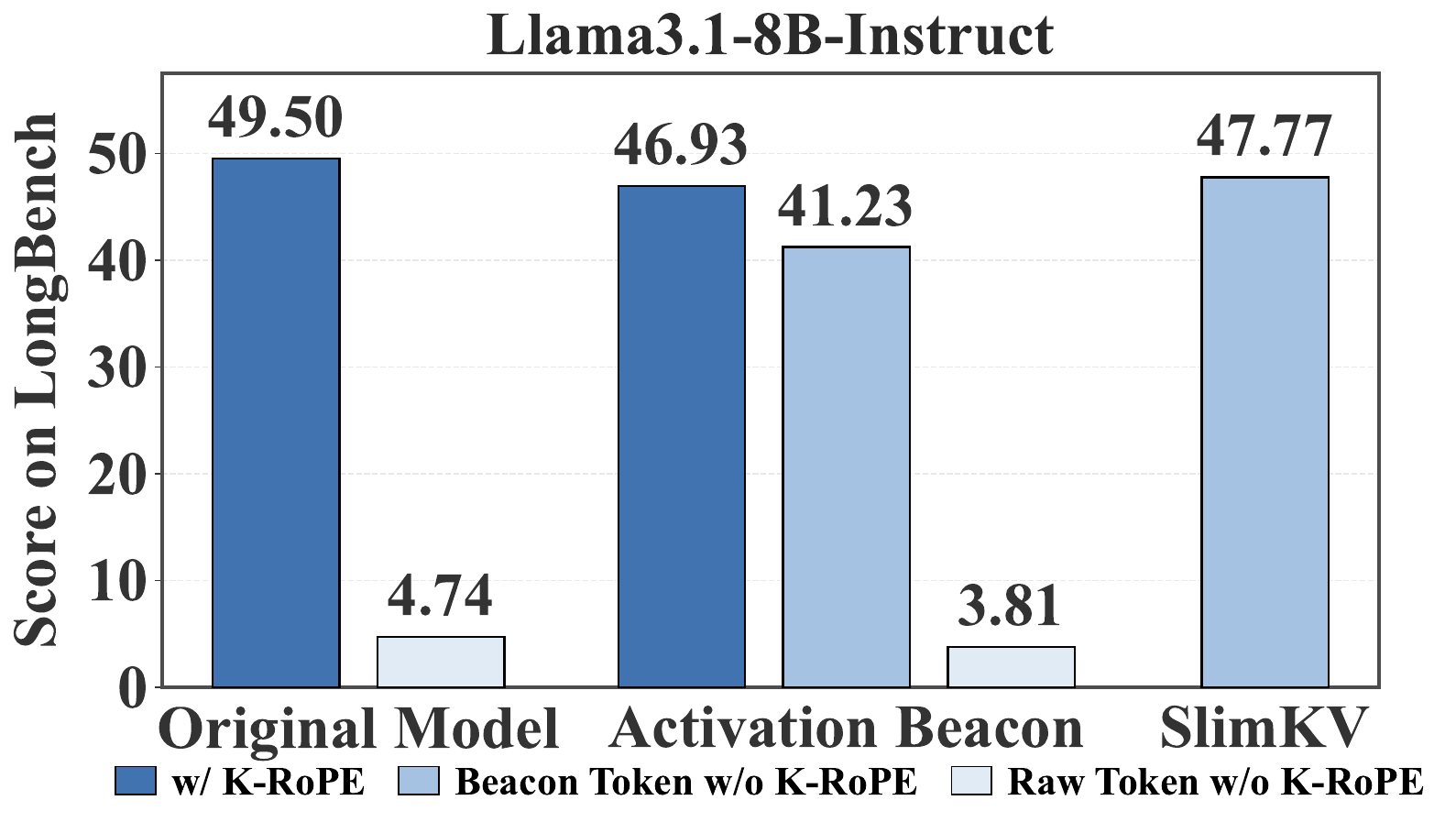}
    \caption{LongBench scores under K-RoPE ablations, showing positional asymmetry between raw and beacon tokens. SlimKV remains close to the original model without beacon K-RoPE.}
    \label{fig:K_RoPE_ablation}
\end{figure}

These latent representations reduce KV-cache storage, but the reconstruction latency bottleneck remains in RoPE-based LLMs~\citep{su2024roformer}. Motivated by the intuition that compressing multiple tokens into a beacon state may preserve partial positional cues in its latent representation, we investigate the role of key-side RoPE and find an empirical positional asymmetry between beacon and raw-token Keys: removing key-side RoPE affects both types of Keys, yet beacon memory tokens exhibit much smaller performance degradation than raw tokens (Figure~\ref{fig:K_RoPE_ablation}). Exploiting this asymmetry, we train beacon KV projections under a K-RoPE-free constraint, allowing K-RoPE to be removed from beacon Keys at inference while preserving RoPE for raw-token Keys and all Queries, enabling reconstruction-free latent-space attention.

We evaluate SlimKV on LongBench, Needle-in-a-Haystack (NIAH), and long-context decoding latency, covering accuracy, positional robustness, and efficiency.

Our contributions are threefold:

\begin{itemize}[leftmargin=*]
    \item We propose \textbf{SlimKV}, a question-agnostic joint token-feature KV-cache compression framework that learns low-rank beacon KV representations and allocates layer-adaptive ranks under a fixed feature budget.

    \item We identify an asymmetric dependence on key-side RoPE between beacon and raw-token Keys, and strengthen this property to enable reconstruction-free latent-space attention by removing K-RoPE only from beacon Keys.

    \item Experiments show that SlimKV achieves clear advantages under high compression, remains leading at lower compression ratios while retaining over 96\% of the uncompressed model's LongBench score, and achieves up to 7.34$\times$ attention speedup and 3.38$\times$ end-to-end decoding speedup.
\end{itemize}

\begin{figure*}[t]
    \centering
    \includegraphics[width=1\linewidth]{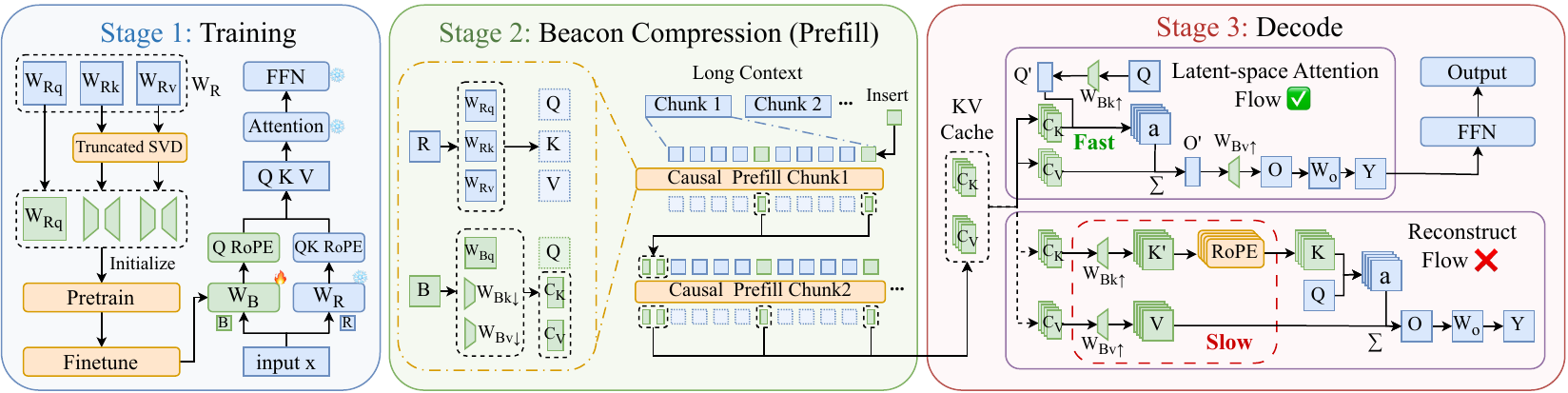}
    \caption{Overview of SlimKV. SlimKV trains low-rank beacon KV projections and caches only latent beacon KV states after prefill. During decoding, SlimKV performs latent-space attention over cached beacon memories, avoiding full-dimensional reconstruction.}
    \label{fig:Overview}
\end{figure*}

\section{Related Work}

\paragraph{Token-wise Compression}
Token-wise KV-cache compression reduces cache size by decreasing the number of historical states used in attention. Existing methods include \emph{prompt-level} and \emph{KV-level} compression. \emph{Prompt-level} methods operate on the input context: hard prompt compression~\citep{jiang-2023-llmlingua,jiang-2024-longllmlingua} removes selected tokens, while soft prompt compression~\citep{mu-2023-GIST,ge-2024-ICAE,li-2025-500xcompressor,zhang-2025-ActivationBeacon} condenses long prompts into learned memory tokens. \emph{KV-level} methods compress generated caches through KV eviction~\citep{zhang-2023-h2o,li2024snapkv,kim2025kvzip} or KV merging~\citep{zhang2024cam,wang2024KVMerger}. However, deletion-based methods may discard useful information, and question-aware variants are less suitable when future queries are unavailable. Soft compression and KV merging avoid explicit deletion but still face capacity bottlenecks under aggressive compression; KV merging also often relies on heuristic aggregation. Moreover, KV-level compression reduces decoding-time cache cost but not prefill computation for the original long prompt.

\paragraph{Feature-wise Compression}
Feature-wise KV-cache compression reduces per-token cache size by shrinking feature representations or removing head redundancy. MQA~\citep{shazeer2019MQA} and GQA~\citep{ainslie2023gqa} share or group K/V heads, while latent-KV architectures such as MLA~\citep{liu2024deepseek} store K/V in a low-dimensional latent space with decoupled positional handling. However, MLA-style designs typically require pre-training from scratch and are less applicable to existing GQA-based backbones. Post-training methods such as Eigen Attention~\citep{saxena-etal-2024-eigen}, Palu~\citep{chang2025palu}, and MHA2MLA~\citep{ji-etal-2025-MHA2MLA} retrofit existing models with low-rank latent KV structures. Nevertheless, post-hoc SVD-style decomposition may introduce information bottlenecks, and low-rank Keys often require full-space reconstruction before RoPE and attention scoring, incurring additional latency.

% \paragraph{Positioning Our Work}
% To our knowledge, SlimKV is the first question-agnostic framework that couples memory-token soft compression with low-rank latent KV compression. It compresses beacon memory along the feature dimension through training-time low-rank learning, and mitigates the reconstruction-latency bottleneck by enabling reconstruction-free latent-space attention during decoding.

\section{Joint Token-Feature Compression}
\subsection{Overview}

SlimKV is a joint token-feature KV-cache compression framework built on the Activation Beacon-style progressive token compression pipeline~\citep{zhang-2025-ActivationBeacon}. The input is chunked with beacon tokens inserted at fixed intervals; each chunk attends to previous beacon memories, allowing beacon states to absorb contextual information. After each chunk-wise prefill, raw-token KV states are discarded and only compressed beacon memories are cached.

SlimKV further extends this pipeline to feature-wise compression by training beacon KV projections directly in their target low-rank form, rather than applying post-hoc decomposition. It caches only latent beacon Keys and Values and allocates layer-adaptive ranks under a global compression budget to match layer-wise heterogeneity. Figure~\ref{fig:Overview} provides an overview.

\subsection{Low-Rank Beacon Training}

SlimKV applies low-rank parameterization only to the beacon branch, with the backbone frozen and only beacon projection parameters updated during training. Let $h_b^{(l)}$ denote a beacon hidden state at layer $l$, and let $D$ be the original KV dimension. Instead of generating full-dimensional beacon Keys and Values, SlimKV first maps beacon states into a layer-specific latent space:
\begin{equation}
    C_K^{(l)} = h_b^{(l)} W_{K,\downarrow}^{(l)}, 
    \quad
    C_V^{(l)} = h_b^{(l)} W_{V,\downarrow}^{(l)},
    \end{equation}
    where both $C_K^{(l)}$ and $C_V^{(l)}$ have dimensions lower than $D$. The corresponding full-dimensional forms for head $m$ can be written as
    \begin{equation}
    K_{b}^{(l)} = C_K^{(l)} U_{K}^{(l)\top},
    \quad
    V_{b}^{(l)} = C_V^{(l)} U_{V}^{(l)\top}.
\end{equation}
During inference, only the latent states $C_K^{(l)}$ and $C_V^{(l)}$ are stored in the KV cache.

Unlike post-training low-rank compression, SlimKV trains beacon KV projections under the target rank constraint from the beginning, allowing beacon memory to adapt to the target latent dimension. Training follows the standard causal language modeling objective over raw tokens. For a raw-token position $t$, the visible context is
\begin{equation}
    \mathcal{C}_t =
    \left\{
    \mathcal{B}_{\mathrm{prev}},
    \mathcal{B}_{<t,\mathrm{cur}},
    \mathcal{R}_{<t,\mathrm{cur}}
    \right\},
\end{equation}
where $\mathcal{B}_{\mathrm{prev}}$ denotes cached latent beacon memory from previous chunks, $\mathcal{B}_{<t,\mathrm{cur}}$ denotes latent beacon memory from the current-chunk prefix, and $\mathcal{R}_{<t,\mathrm{cur}}$ denotes raw-token states from the current-chunk prefix. The training objective is
\begin{equation}
    \mathcal{L}
    =
    -\sum_{t \in \mathcal{T}_{\mathrm{raw}}}
    \log p_\theta
    \left(
    x_t \mid \mathcal{C}_t
    \right),
\end{equation}
where $\mathcal{T}_{\mathrm{raw}}$ denotes non-beacon positions and $\theta$ denotes model parameters. Beacon tokens are excluded from the loss.

\begin{figure}
    \centering
    \includegraphics[width=1\linewidth]{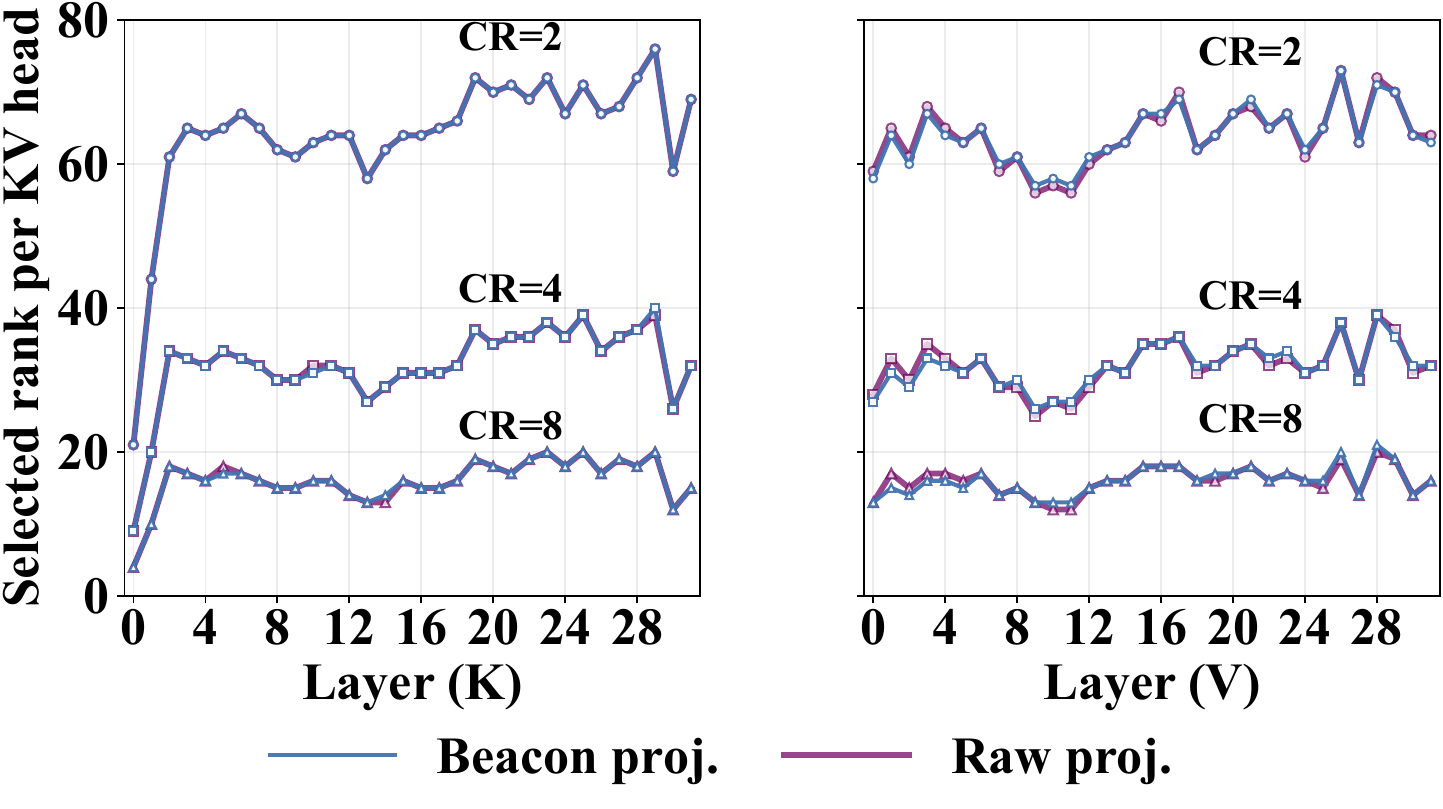}
    \caption{Layer-wise rank allocation. K/V ranks of Beacon and raw weights in Llama-3.1-8B under 2$\times$, 4$\times$, and 8$\times$ feature compression, selected by equal explained variance.}
    \label{fig:Llama_comprasion_layerrank}
\end{figure}

\subsection{Layer-Adaptive Rank Allocation}
\label{sec:rank_allocation}

A uniform latent rank ignores the heterogeneous low-rank structure of Transformer projections. Since trained beacon projections are unavailable before training, SlimKV uses pretrained raw-token KV projection weights as proxies. Their layer-wise rank patterns are highly consistent with those of beacon projection weights, as shown in Figure~\ref{fig:Llama_comprasion_layerrank} and Table~\ref{tab:svd-rank-consistency}. SlimKV therefore derives beacon ranks from raw-token singular-value spectra and initializes each low-rank beacon projection with the truncated SVD of its raw-token counterpart.

We use a floor-constrained equal-energy criterion. Let $\sigma_{l,1} \ge \cdots \ge \sigma_{l,D}$ denote the singular values of the pretrained raw-token KV projection at layer $l$, and define the cumulative spectral energy as
\begin{equation}
    E_l(r)=
    \frac{\sum_{j=1}^{r}\sigma_{l,j}^{2}}
    {\sum_{j=1}^{D}\sigma_{l,j}^{2}} .
\end{equation}
Given an energy threshold $\rho$, the initial rank is
\begin{equation}
    r_l^{0} = \min \{ r \mid E_l(r) \ge \rho \}.
\end{equation}

Pure equal-energy allocation may assign very small ranks to layers with concentrated spectra, creating training bottlenecks. We therefore impose a rank floor. Given a total rank budget $B$ and $L$ layers, where $B=LD/k$ for feature compression ratio $k$, let $\bar r=B/L$ be the average rank and set
\begin{equation}
    r_{\min}=\lceil \lambda \bar r \rceil .
\end{equation}
where $\lambda$ is a hyperparameter that controls the minimum rank. The final rank is
\begin{equation}
    r_l(\rho)=\max(r_l^{0}, r_{\min}).
\end{equation}
We choose the largest $\rho$ that satisfies the global budget:
\begin{equation}
    \rho^\star = \max\{\rho \mid \sum_{l=1}^{L} r_l(\rho) \le B\}.
\end{equation}
Any unused rank budget after integer assignment is allocated to layers with the largest marginal energy gain. This preserves balanced spectral energy, avoids under-allocation, and fixes all beacon ranks.

\begin{figure}[t]
    \centering
    \includegraphics[width=1\linewidth]{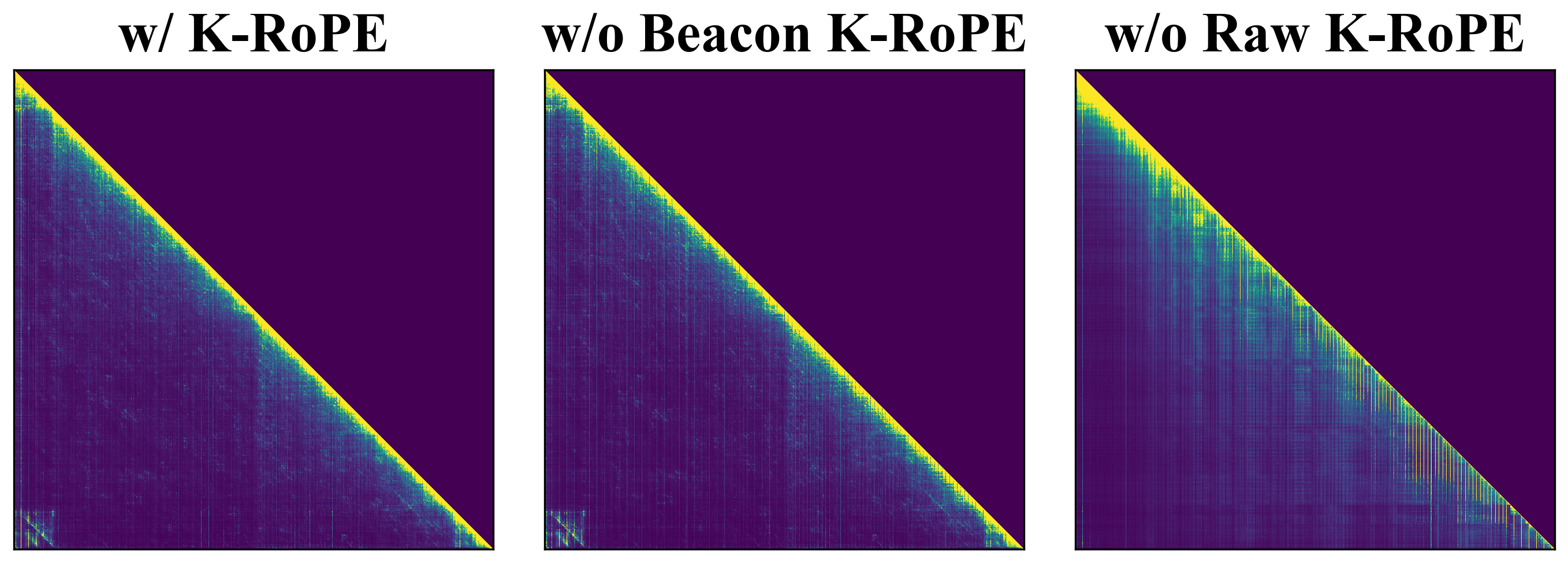}
    \caption{Attention maps under K-RoPE ablations. Beacon K-RoPE removal remains much closer to the K-RoPE-kept pattern than raw-token K-RoPE removal.}
    \label{fig:attn_heatmap}
\end{figure}

\begin{table}[t]
    \centering
    \small
    \setlength{\tabcolsep}{4pt}
    \caption{Similarity to the w/ K-RoPE setting.}
    \label{tab:heatmap_similarity}
    \resizebox{0.92\columnwidth}{!}{%
    \begin{tabular}{lcccc}
        \toprule
        \textbf{Variant} & \textbf{Attn.} $\uparrow$ & \textbf{Block} $\uparrow$ & \textbf{Lag JS} $\downarrow$ & \textbf{Col. JS} $\downarrow$ \\
        \midrule
        Beacon removed & 0.9707 & 0.9982 & 0.0002 & 0.0030 \\
        Raw removed    & 0.7570 & 0.8922 & 0.0062 & 0.0349 \\
        \bottomrule
    \end{tabular}
    }
\end{table}

\section{From Compression to Acceleration: Latent-Space Attention}
\label{sec:latent_attention}

The previous section reduces KV-cache size, but Key reconstruction can still limit speed. Can compressed beacon Keys enable latent-space decoding without full-Key reconstruction?

\subsection{RoPE-aware Beacon Generation}

Although cached beacon Keys are RoPE-free, beacon Queries still retain RoPE and attend to RoPE-applied raw-token Keys during beacon generation. Thus, beacon states are formed through RoPE-aware QK interactions~\citep{su2024roformer}:
\[
\alpha_{b,i}
=
\operatorname{softmax}_i
\left(
\mathrm{RoPE}(q_b,p_b)^\top
\mathrm{RoPE}(k_i,p_i)
\right),
\]
\begin{equation}
h_b
=
\sum_i \alpha_{b,i} v_i .
\end{equation}
Here, $\alpha_{b,i}$ is the attention from beacon token $b$ to raw token $i$, and $p_b,p_i$ denote their RoPE positions. Since beacon states are formed through RoPE-aware QK interactions, they can absorb coarse positional cues from raw tokens even though explicit K-RoPE is removed during beacon generation.

\subsection{Positional Asymmetry}

SlimKV leverages the positional asymmetry between raw and beacon tokens. Raw tokens are lexical units aligned with token-level RoPE, whereas beacon tokens aggregate span-level context and cannot be faithfully tied to a single RoPE index. K-side RoPE ablations in Figure~\ref{fig:K_RoPE_ablation} show that removing K-RoPE degrades both raw-token and beacon Keys, but the drop is much smaller for beacon Keys. Attention maps corroborate this pattern (Figure~\ref{fig:attn_heatmap}): beacon K-RoPE removal largely preserves the diagonal structure, while raw-token K-RoPE removal distorts it. Table~\ref{tab:heatmap_similarity} quantifies this observation, with metric definitions provided in Appendix~\ref{app:attention_similarity_metrics}. Figure~\ref{fig:context_reconstruction} further shows that RoPE-free beacon states reconstruct random character sequences far above random guessing, indicating non-trivial positional cues without explicit K-RoPE.

\begin{figure}[t]
    \centering
    \includegraphics[width=1.0\linewidth]{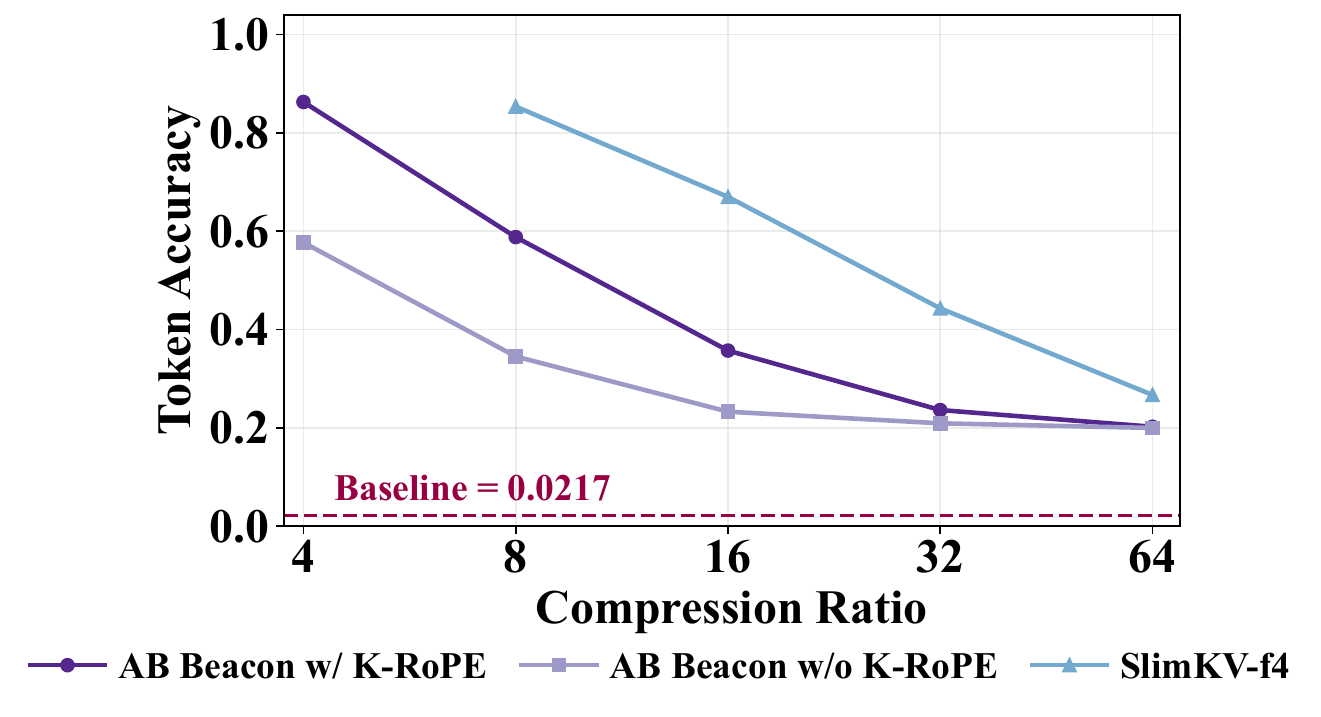}
    \caption{
    Reconstruction accuracy from compressed beacon states on length-16,384 random sequences over 46 symbols, averaged over three seeds. SlimKV starts from 8$\times$ total compression due to 4$\times$ feature compression. Dashed line denotes random-guessing probability.
    }
    \label{fig:context_reconstruction}
\end{figure}

\subsection{Reconstruction-free Latent-Space Attention}

Based on these findings, SlimKV imposes a RoPE-free constraint on beacon Keys during training, encouraging beacon states to acquire implicit positional cues during compression. With explicit beacon K-RoPE removed, SlimKV performs latent-space attention (LSA) during decoding to avoid full-Key reconstruction. For one decoding step, let $C_K,C_V\in\mathbb{R}^{N_b\times D/k}$ denote cached latent beacon K/V states, with up-projections $U_K,U_V$. If beacon Keys retain K-RoPE, attention requires
\begin{equation}
    Q' \operatorname{RoPE}(K_b)^\top
    =
    Q' \operatorname{RoPE}(C_KU_K^\top)^\top ,
\end{equation}
where $Q'$ denotes RoPE-applied Queries. RoPE is position-dependent and cannot be absorbed into a single position-independent projection. Removing beacon K-RoPE eliminates this obstacle and allows attention to be associated in the latent space:
\begin{equation}
    Q'K_b^\top
    =
    Q'(C_KU_K^\top)^\top
    =
    (Q'U_K)C_K^\top ,
\end{equation}
where $Q'U_K$ is computed after GQA grouping. The Value path is handled analogously:
\begin{equation}
    AV_b
    =
    A(C_VU_V^\top)
    =
    (AC_V)U_V^\top .
\end{equation}
Thus, LSA computes attention scores and aggregates Values directly in the latent space.

\subsection{Cost Analysis}
\label{sec:cost_analysis}

We analyze the dominant FLOPs per decoding step, deferring the comprehensive derivation to Appendix~\ref{app:cost_derivation}. Let $H_q$ and $H_{\mathrm{kv}}$ denote the number of query and KV heads, $g = H_q / H_{\mathrm{kv}}$ the GQA group size, $d$ the per-head dimension, $D = H_{\mathrm{kv}}d$ the total KV dimension, $k$ the feature compression ratio, and $N_b$ the number of cached beacon tokens. 

\begin{table*}[t]
    \centering
    \small
    \setlength{\tabcolsep}{4.0pt}
    \renewcommand{\arraystretch}{1.12}
    \caption{
    LongBench scores under high KV-cache compression. SlimKV \texttt{t$n$f$m$}: $n\times$ token-wise, $m\times$ feature-wise. The \textbf{bold} and \underline{underlined} numbers denote the first and second rankings, respectively.
    }
    \vspace{-0.5em}
    \resizebox{0.98\textwidth}{!}{%
    \begin{tabular}{cc@{\hspace{1.1em}}cccccc@{\hspace{1.5em}}cccccc}
        \toprule
        \multirow{2}{*}{\textbf{Comp.}} & \multirow{2}{*}{\textbf{Method}}
        & \multicolumn{6}{c}{\textsc{Llama-3.1-8B}}
        & \multicolumn{6}{c}{\textsc{Qwen2.5-14B}} \\
        \cmidrule(lr){3-8} \cmidrule(lr){9-14}
        & & \textbf{S-Doc} & \textbf{M-Doc} & \textbf{Summ.}
        & \textbf{Few} & \textbf{Code} & \textbf{Avg.}
        & \textbf{S-Doc} & \textbf{M-Doc} & \textbf{Summ.}
        & \textbf{Few} & \textbf{Code} & \textbf{Avg.} \\
        \midrule
        1$\times$ & Full KV
        & 44.34 & 46.65 & 28.71 & 69.45 & 58.37 & 49.50
        & 42.72 & 43.73 & 28.05 & 72.04 & 61.05 & 51.52 \\
        \midrule
        \multirow{8}{*}{16$\times$} & KVZip
        & 33.86 & 29.11 & 23.97 & 22.97 & 34.89 & 28.96
        & 7.83 & 5.48 & 3.27 & 35.12 & 14.20 & 13.18 \\
        & PALU
        & 13.18 & 9.80 & 12.75 & 37.86 & 21.04 & 18.92
        & 17.77 & 14.18 & 21.40 & 59.40 & 25.75 & 27.70 \\
        & SnapKV
        & \underline{38.55} & \textbf{43.30} & 23.61 & \textbf{65.92} & 56.62 & 45.60
        & 36.74 & \textbf{51.20} & 20.22 & \underline{68.50} & 52.37 & 45.81 \\
        & Activation Beacon
        & 34.72 & 36.28 & 25.34 & 65.01 & \underline{65.14} & 45.30
        & 35.76 & 38.31 & 25.24 & 63.66 & \textbf{57.77} & 44.15 \\
        & SlimKV-f2~(\texttt{t8f2})
        & 37.64 & \underline{37.40} & \underline{26.25} & 65.11 & 64.55 & \underline{46.19}
        & \textbf{39.44} & 42.16 & \textbf{25.91} & 67.20 & \underline{57.72} & \textbf{46.48} \\
        & SlimKV-f4~(\texttt{t4f4})
        & \textbf{38.65} & 34.33 & \textbf{26.57} & 65.14 & \textbf{67.11} & \textbf{46.36}
        & 38.15 & \underline{43.76} & 25.50 & \textbf{69.16} & 54.61 & \underline{46.23} \\
        & SlimKV-f8~(\texttt{t2f8})
        & 33.46 & 36.24 & 26.02 & \underline{65.68} & 63.82 & 45.05
        & \underline{38.25} & 38.83 & \underline{25.61} & 68.06 & 50.89 & 44.33 \\
        \midrule
        \multirow{8}{*}{32$\times$} & KVZip
        & 9.50 & 3.35 & 12.69 & 0.96 & 20.48 & 9.40
        & 1.09 & 0.78 & 1.32 & 1.53 & 20.91 & 5.12 \\
        & PALU
        & 1.58 & 0.56 & 5.47 & 1.32 & 15.02 & 4.79
        & 4.20 & 2.82 & 10.90 & 14.66 & 13.67 & 9.25 \\
        & SnapKV
        & \textbf{35.80} & \textbf{42.82} & 21.61 & 62.70 & 53.29 & 43.24
        & 32.45 & \textbf{49.61} & 18.09 & 62.71 & 48.54 & 42.28 \\
        & Activation Beacon
        & 29.94 & 35.24 & 24.53 & 64.26 & 64.35 & 43.67
        & 32.80 & 35.64 & 25.16 & 63.20 & 57.01 & 42.76 \\
        & SlimKV-f2~(\texttt{t16f2})
        & 33.49 & 35.34 & 24.86 & 61.19 & 59.68 & 42.91
        & 37.15 & 38.59 & 25.49 & 64.97 & \textbf{57.09} & 44.66 \\
        & SlimKV-f4~(\texttt{t8f4})
        & \underline{34.16} & \underline{35.75} & \textbf{25.43} & 63.35 & \textbf{65.07} & \textbf{44.75}
        & \textbf{37.55} & \underline{40.29} & \textbf{25.50} & \textbf{68.76} & 53.36 & \textbf{45.09} \\
        & SlimKV-f8~(\texttt{t4f8})
        & 32.53 & 35.41 & \underline{25.14} & \textbf{64.36} & \underline{62.22} & \underline{43.93}
        & \underline{35.83} & 37.99 & \underline{25.27} & \underline{68.27} & \underline{56.10} & \underline{44.69} \\
        \bottomrule
    \end{tabular}
    }
    \label{tab:longbench_main}
\end{table*}

\begin{table}[t]
    \centering
    \normalsize
    \setlength{\tabcolsep}{5pt}
    \caption{LongBench gap reduction over Activation Beacon. $\Delta$ denotes the drop from Full KV.}
    \renewcommand{\arraystretch}{1.03}
    \vspace{-0.5em}
    \resizebox{\columnwidth}{!}{%
    \begin{tabular}{cccccc}
        \toprule
        \textbf{Backbone} & \textbf{Comp.}
        & $\Delta_{\textbf{AB}}$ 
        & $\Delta_{\textbf{SlimKV-f4}}$
        & $\Delta$ \textbf{Red.}
        & \textbf{Gap Red.} \\
        \midrule
        \multirow{2}{*}{\textsc{Llama-3.1-8B}}
        & 16$\times$ & 4.20 & 3.14 & 1.06 & 25.2\% \\
        & 32$\times$ & 5.83 & 4.75 & 1.08 & 18.5\% \\
        \midrule
        \multirow{2}{*}{\textsc{Qwen2.5-14B}}
        & 16$\times$ & 7.37 & 5.29 & 2.08 & 28.2\% \\
        & 32$\times$ & 8.76 & 6.43 & 2.33 & 26.6\% \\
        \bottomrule
    \end{tabular}
    }
    \label{tab:gap_reduction}
\end{table}

The reconstruction-based baseline with beacon K-RoPE has arithmetic cost:
\begin{equation}
    F_{\mathrm{recon}} \approx N_b \frac{2D^2}{k} + (1+2g)N_b D ,
\end{equation}
where the first and second terms represent full KV reconstruction and standard attention computation, respectively. In contrast, SlimKV incurs a cost of:
\begin{equation}
    F_{\mathrm{SlimKV}} \approx \frac{2gD^2}{k} + \frac{2H_q}{k}N_b D ,
\end{equation}
where the first and second terms correspond to per-step query/output projection overhead and latent-space attention computation.

Although both methods require $\mathcal{O}(N_bD)$ attention computation, the reconstruction baseline additionally materializes full-dimensional KV states at $\mathcal{O}(N_bD^2/k)$ cost. This sequence-dependent term can dominate long-context decoding, while SlimKV replaces it with an $\mathcal{O}(gD^2/k)$ per-step projection cost independent of $N_b$. This analysis covers arithmetic cost only, with empirical latency reported in Section~\ref{sec:efficiency}.

During decoding, newly generated tokens are cached as raw tokens. This leads to a hybrid attention scheme where beacon queries use latent-space attention and raw-token queries use standard attention, as detailed in Appendix~\ref{hybrid_attention}. Since beacon states dominate long-context caches, optimizing beacon-query latency remains important.

\section{Experiments}

\vspace{-0.3em}
\subsection{Experimental Setup}

\begin{figure*}[t]
    \centering

    \begin{subfigure}[t]{\linewidth}
        \centering
        \includegraphics[width=0.92\linewidth]{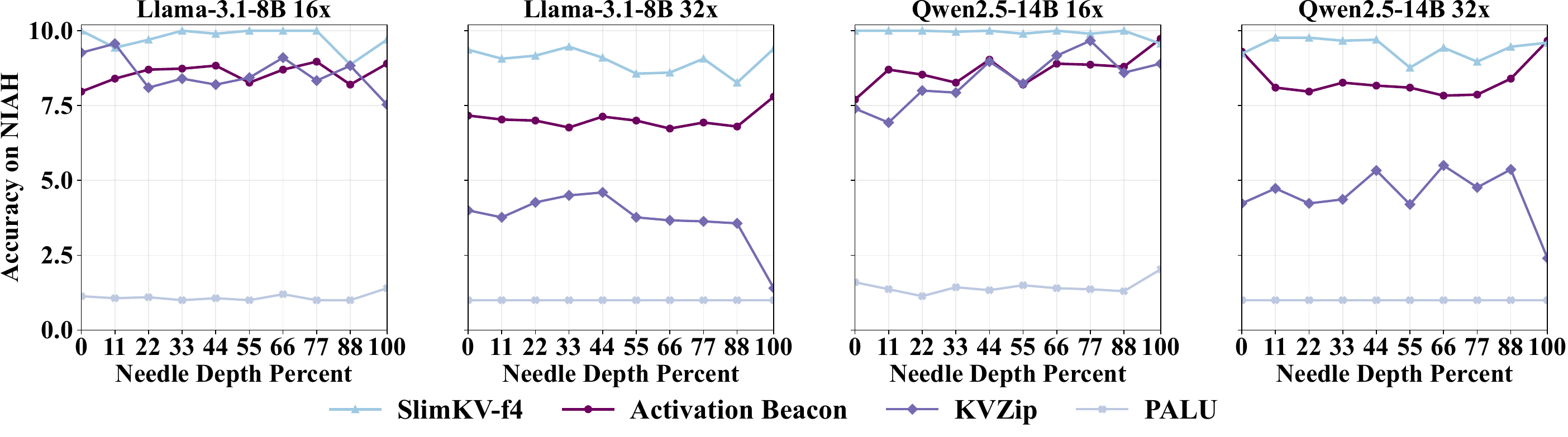}
        \caption{}
        \label{fig:NIAH_line}
    \end{subfigure}

    \begin{subfigure}[t]{\linewidth}
        \centering
        \includegraphics[width=0.92\linewidth]{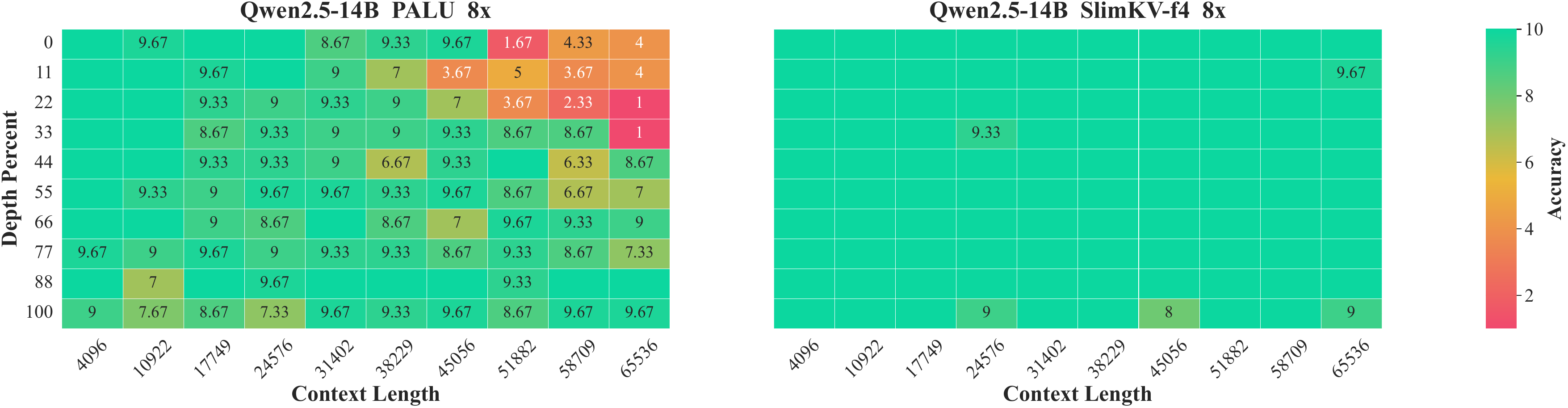}
        \vspace{-0.3em}
        \caption{}
        \label{fig:Qwen25_r8_NIAH}
    \end{subfigure}
    \vspace{-0.3em}
    \caption{Needle-in-a-Haystack evaluation. (a) SlimKV outperforms baselines across depths under 16$\times$ and 32$\times$ compression. (b) SlimKV maintains retrieval ability for information beyond Qwen2.5-14B's context window (32K). Since higher ratios limit PALU's retrieval accuracy, we select 8$\times$ compression to better reveal the behavior.}
    \label{fig:NIAH_combined}
    \vspace{-0.4em}
\end{figure*}

\paragraph{Models and training.}
We evaluate SlimKV on Llama-3.1-8B-Instruct~\citep{grattafiori2024llama3herdmodels} and Qwen2.5-14B-Instruct~\citep{qwen2.5, qwen2}. Backbones are frozen, and training is restricted to the beacon-branch QKV projections. We follow a two-stage recipe: pre-training on 0.5B RedPajama~\citep{weber2024redpajama} tokens, followed by fine-tuning on LongAlpaca~\citep{chen2024longlora}, BookSum~\citep{kryscinski-etal-2022-booksum}, and GPT-generated synthetic data~\citep{zhang-2025-ActivationBeacon}. All samples are shorter than 20K tokens. We use batch size 8, learning rates of $5\times10^{-5}$/$1\times10^{-5}$, linear decay, and no warmup. We apply the auto-regressive loss only to non-beacon tokens, while beacon tokens serve as compressed memory states and are excluded from supervision.

\vspace{-0.5em}
\paragraph{Compression settings.}
SlimKV compresses KV cache along token and feature dimensions. Token compression is controlled by the beacon interval, and feature compression by beacon KV latent ranks, with $2\times$, $4\times$, and $8\times$ settings evaluated. Raw-token Keys and all Queries retain RoPE, while beacon Keys remove RoPE. Layer-wise ranks follow the floor-constrained equal-energy criterion in Section~\ref{sec:rank_allocation}, with low-rank beacon projections initialized from truncated SVD of raw-token KV projections. We set $\lambda=0.5$ to balance global rank budget and minimum layer capacity.
\vspace{-0.5em}
\paragraph{Baselines.}
We compare SlimKV with Full KV and representative methods: PALU~\citep{chang2025palu} for low-rank KV compression, KVZip~\citep{kim2025kvzip} and SnapKV~\citep{li2024snapkv} for KV eviction, and Activation Beacon~\citep{zhang-2025-ActivationBeacon} for prompt-based soft compression. SlimKV and Activation Beacon require offline training, whereas PALU, KVZip, and SnapKV are training-free. For trainable methods, we use the same training data for fair comparison.

\begin{table}[t]
    \centering
    \normalsize
    \setlength{\tabcolsep}{5pt}
    \caption{LongBench averages at low compression. The averages of the uncompressed models are in Table~\ref{tab:longbench_main}.}
    \vspace{-0.5em}
    \renewcommand{\arraystretch}{0.9}
    \renewcommand{\arraystretch}{1}
    \resizebox{\columnwidth}{!}{%
    \begin{tabular}{cccccc}
        \toprule
        \textbf{Backbone} & \textbf{Comp.}
        & \textbf{KVZip} & \textbf{PALU}
        & \textbf{AB} & \textbf{SlimKV-f2} \\
        \midrule
        \multirow{2}{*}{\textsc{Llama-3.1-8B}}
        & 4$\times$ & 36.39 & 46.66 & \textbf{48.21} & \smash{\underline{48.10}} \\
        & 8$\times$ & 37.25 & 46.42 & \smash{\underline{46.93}}  & \textbf{47.77} \\
        \midrule
        \multirow{2}{*}{\textsc{Qwen2.5-14B}}
        & 4$\times$ & 47.72 & 45.95 & \smash{\underline{48.86}} & \textbf{49.92} \\
        & 8$\times$ & \smash{\underline{46.74}} & 45.15 & 45.50 & \textbf{49.94} \\
        \bottomrule
    \end{tabular}
    }
    \label{tab:longbench_low_ratio}
\end{table}
\vspace{-0.5em}
\subsection{Effectiveness Evaluation}

We evaluate SlimKV on LongBench~\citep{bai-etal-2024-longbench} and Needle-in-a-Haystack (NIAH)~\citep{kamradt2023needle}. LongBench measures long-context performance across QA, summarization, few-shot learning, and code, while NIAH probes position-sensitive retrieval. We mainly study 16$\times$ and 32$\times$ compression, and additionally report 4$\times$ and 8$\times$ LongBench results to assess robustness across compression ratios.

\paragraph{LongBench.}
Table~\ref{tab:longbench_main} reports the LongBench results. Across both backbones and high compression ratios, SlimKV achieves the best average score. Table~\ref{tab:gap_reduction} further shows that SlimKV-f4 reduces the gap to the uncompressed model by 18.5\%--28.2\% over Activation Beacon, demonstrating its advantage over token-only soft compression. Comparing 2$\times$, 4$\times$, and 8$\times$ feature compression under the same overall compression ratio reveals a trade-off: too little feature compression forces aggressive token compression, while too much feature compression restricts the latent KV capacity. The 4$\times$ setting strikes the best balance and achieves the highest average score in most high-compression settings.

At lower 4$\times$ and 8$\times$ compression ratios, SlimKV uses 2$\times$ feature compression to retain non-trivial token compression. As shown in Table~\ref{tab:longbench_low_ratio}, SlimKV preserves over 96\% of the uncompressed average score, consistently outperforms KVZip and PALU, remains competitive with Activation Beacon, and achieves the best score in most low-ratio settings. This highlights SlimKV's robustness across different compression ratios.

\paragraph{Question-Aware vs. Question-Agnostic.}
In our LongBench evaluation, we include SnapKV as a question-aware baseline to contrast with SlimKV's question-agnostic paradigm and position SlimKV within the broader KV-cache compression landscape. SnapKV leverages the known question for targeted KV selection~\citep{li2024snapkv}, whereas SlimKV is question-agnostic and does not rely on the question for compression. SnapKV is well suited to single-request, question-specific retrieval, while SlimKV better supports multi-turn and multi-request scenarios. Despite SnapKV's question-aware advantage, SlimKV achieves higher LongBench averages at 16$\times$ and 32$\times$ compression.

To highlight this distinction, we construct a multi-request setting on QMSum~\citep{zhong2021qmsum}, where each context is compressed once and reused for multiple unseen questions. Full KV achieves 25.17 ROUGE-L F1, while at 16$\times$ compression, SnapKV drops to 10.39 and SlimKV-f4 reaches 24.47, retaining 97.2\% of Full KV performance. This suggests an advantage of SlimKV in multi-request scenarios.

\paragraph{Needle-in-a-Haystack (NIAH).}

We further evaluate position-sensitive retrieval with NIAH, which is particularly relevant because removing K-RoPE from beacon may raise concerns about positional robustness~\citep{liu2024lost}. NIAH results are averaged over three random seeds, using GPT-5.4-mini~\citep{openai2026gpt54mini} as the judge. Figure~\ref{fig:NIAH_line} shows SlimKV-f4 outperforms baselines across backbones and compression ratios. Its stable performance across needle depths indicates robust retrieval under varying positions.

As shown in Figure~\ref{fig:Qwen25_r8_NIAH}, when the target information lies beyond Qwen2.5-14B's context window, PALU shows reduced accuracy, whereas SlimKV maintains stable performance across context lengths and needle depths. This is partly because SlimKV compresses long context into fewer beacon tokens, keeping the cache within the usable context range. Meanwhile, the stable retrieval results show that K-RoPE-free beacon memories remain effective for beyond-window information.

\begin{table}[t]
    \centering
    \normalsize
    \setlength{\tabcolsep}{3.5pt}
    \renewcommand{\arraystretch}{1.18}
    \caption{LongBench category scores on Qwen3-30B-A3B-Instruct-2507.}
    \vspace{-0.5em}
    \resizebox{\columnwidth}{!}{%
    \begin{tabular}{lcccccccc}
        \toprule
        \textbf{Method} & \textbf{Comp.} & \textbf{S-Doc} & \textbf{M-Doc} & \textbf{Summ.} & \textbf{Few-shot} & \textbf{Code} & \textbf{Avg.} \\
        \midrule
        Full KV   & 1$\times$  & 42.87 & 51.02 & 25.60 & 71.26 & 60.75 & 50.30 \\
        SlimKV-f4 & 16$\times$ & 36.87 & 40.43 & 26.00 & 68.24 & 58.87 & 46.08 \\
        SlimKV-f4 & 32$\times$ & 34.14 & 38.16 & 24.77 & 67.85 & 63.09 & 45.60 \\
        \bottomrule
    \end{tabular}
    }
    \label{tab:qwen3_moe}
\end{table}

\subsection{Scalability and Parameter Analysis}
\label{sec:scalability}

To assess scalability beyond the dense backbones used in the main evaluation, we evaluate Qwen3-30B-A3B-Instruct-2507~\citep{yang2025qwen3}. We compare the uncompressed model with SlimKV-f4 at 16$\times$ and 32$\times$ compression, which evaluates SlimKV at a larger scale and on an MoE LLM. 

Table~\ref{tab:qwen3_moe} shows that SlimKV-f4 preserves competitive LongBench performance with only 434M trainable parameters, which account for 1.45\% of the original model size. Because trainable-parameter storage scales with the hidden dimension, SlimKV-f4 uses fewer trainable parameters on this 30B MoE backbone than on Llama-3.1-8B and Qwen2.5-14B, which require 621M and 1.41B trainable parameters, respectively. These results demonstrate parameter-efficient scalability and preliminary transfer to larger MoE LLMs.

\vspace{-0.5em}
\subsection{Efficiency Evaluation}
\label{sec:efficiency}

We evaluate efficiency on Llama-3.1-8B at 16$\times$ and 32$\times$ compression, corresponding to \texttt{t4f4} and \texttt{t8f4}, by comparing SlimKV with a reconstruction-based variant that retains beacon K-RoPE and materializes full-dimensional K/V states before attention, with the uncompressed model as reference. We mainly report single-step attention latency and end-to-end per-token decoding latency.

Experiments use 32K--128K contexts and generate 128 tokens per run for average per-token latency. All methods use batch size 1, BF16 precision, and PyTorch eager execution on a single A100 80GB GPU. We report the median over three runs after one warm-up. Synthetic full KV-cache tensors are used for the uncompressed model to avoid prefill memory limits. Speedups are reported in Table~\ref{tab:decode_speedup}.

Due to OOM at longer contexts, the uncompressed model is evaluated for Time-to-First-Token (TTFT) and peak memory at its longest feasible length of 11.5K tokens. At this length, SlimKV \texttt{t4f4}/\texttt{t8f4} achieve 1.03$\times$/1.37$\times$ TTFT speedups over the uncompressed model and reduce peak memory by 2.71$\times$/2.82$\times$, respectively. Longer-context results are provided in Appendix~\ref{app:efficiency_details}.

\begin{table}[t]
    \centering
    \small
    \setlength{\tabcolsep}{3.8pt}
    \renewcommand{\arraystretch}{1.12}
    \caption{Decoding speedup over the uncompressed model on Llama-3.1-8B-Instruct. Attn./E2E denote attention/per-token latency.}
    \vspace{-0.3em}
    \resizebox{0.95\columnwidth}{!}{%
    \begin{tabular}{ccccccc}
        \toprule
        \textbf{Comp.} & \textbf{Metric} & \textbf{Method}
        & \textbf{32K} & \textbf{64K} & \textbf{96K} & \textbf{128K} \\
        \midrule
        \multirow{4}{*}{16$\times$}
        & \multirow{2}{*}{Attn.}
        & Recon  & 1.76$\times$ & 2.81$\times$ & 3.63$\times$ & 5.34$\times$ \\
        &
        & SlimKV & 1.94$\times$ & 3.81$\times$ & 5.08$\times$ & 7.34$\times$ \\
        \cmidrule(lr){2-7}
        & \multirow{2}{*}{E2E}
        & Recon  & 1.13$\times$ & 1.64$\times$ & 2.13$\times$ & 2.86$\times$ \\
        &
        & SlimKV & 1.22$\times$ & 2.04$\times$ & 2.63$\times$ & 3.38$\times$ \\
        \midrule
        \multirow{4}{*}{32$\times$}
        & \multirow{2}{*}{Attn.}
        & Recon  & 1.52$\times$ & 2.35$\times$ & 3.84$\times$ & 4.92$\times$ \\
        &
        & SlimKV & 1.62$\times$ & 2.54$\times$ & 4.36$\times$ & 6.09$\times$ \\
        \cmidrule(lr){2-7}
        & \multirow{2}{*}{E2E}
        & Recon  & 1.15$\times$ & 1.67$\times$ & 2.60$\times$ & 3.06$\times$ \\
        &
        & SlimKV & 1.23$\times$ & 1.75$\times$ & 2.82$\times$ & 3.49$\times$ \\
        \bottomrule
    \end{tabular}
    }
    \label{tab:decode_speedup}
\end{table}

\subsection{Ablation Studies}

\begin{table}[t]
    \centering
    \small
    \setlength{\tabcolsep}{5.0pt}
    \renewcommand{\arraystretch}{1.12}
    \caption{
    Ablation study on LongBench at 32$\times$ compression with Llama-3.1-8B-Instruct.
    }
    \vspace{-0.3em}
    \resizebox{0.46\textwidth}{!}{%
    \begin{tabular}{cccccc}
        \toprule
        \textbf{T-wise} & \textbf{F-wise} & \textbf{Train-R} & \textbf{Infer-R} & \textbf{Ada-r} & \textbf{Score}\\
        \midrule
        \cmark & \xmark  & \cmark & \cmark  & \xmark  & 43.67 \\
        \midrule
        \cmark & \xmark  & \cmark & \xmark   & \xmark  & 41.28 \\
        \cmark & \xmark  & \xmark & \xmark   & \xmark  & 41.95 \\
        \cmark & \cmark  & \xmark & \xmark   & \xmark  & 44.38 \\
        \cmark & \cmark  & \xmark & \xmark   & \cmark  & 44.75 \\
        \bottomrule
    \end{tabular}
    }
    \label{tab:ablation}
\end{table}

We ablate SlimKV on Llama-3.1-8B-Instruct under 32$\times$ total compression. Token-only variants use 32$\times$ token-wise compression, while joint variants use 8$\times$ token-wise and 4$\times$ feature-wise (\texttt{t8f4}). T-wise/F-wise denote token and feature compression, Train-R/Infer-R denote beacon K-RoPE during training/inference, and Ada-r denotes layer-adaptive ranks.

Removing beacon K-RoPE does degrade accuracy, showing that RoPE removal alone is insufficient. SlimKV instead trains beacon projections directly in a low-rank form under the no-K-RoPE constraint. Layer-adaptive rank allocation further improves performance, and the full design ultimately surpasses Activation Beacon.

\section{Conclusion}

We introduced SlimKV, a question-agnostic joint token-feature KV-cache compression framework for long-context LLMs. SlimKV learns layer-adaptive low-rank beacon KV representations and removes key-side RoPE only from beacon Keys, enabling reconstruction-free latent-space attention. Experiments on LongBench, Needle-in-a-Haystack, and decoding efficiency demonstrate a stronger accuracy-efficiency trade-off than prior KV-cache compression methods.

\section*{Limitations} 

First, SlimKV is mainly evaluated on RoPE-based language models, which represent a widely used family of current LLM backbones. Since our reconstruction-free design addresses the key-side RoPE reconstruction bottleneck in low-rank attention, its behavior under other positional encoding schemes~\citep{raffel2020exploring,press2022train,peng2024yarn} remains to be further examined. 

Second, our experiments focus on text-only long-context language models. Extending SlimKV to broader architectures, such as multimodal LLMs with visual or cross-modal KV caches, is an interesting direction but requires additional validation.
\section*{Acknowledgements}

This work was partially supported by the Ministry of Education, Singapore, under its Academic Research Fund Tier 2 (MOE-T2EP20224-0006), and partially by the National Natural Science Foundation of China under Grant 92473114.
\bibliography{references}

\clearpage
\appendix

\section{Appendix}
\label{sec:appendix}

\subsection{Quantitative Analysis of Layer-wise Rank Consistency}
\label{app:rank_consistency}

We provide additional quantitative evidence for the layer-wise rank consistency between Beacon and raw K/V projections. For each target compression ratio, ranks are selected using the same explained-variance criterion under an identical average rank budget. In Table~\ref{tab:svd-rank-consistency}, Beacon and Raw report the average selected ranks, Spearman measures whether the two rank profiles induce the same layer ordering, and MAE / Max reports the mean and maximum absolute per-layer rank differences.

\begin{table}[h]
    \centering
    \caption{Layer-wise rank consistency between Beacon and raw K/V projections.}
    \label{tab:svd-rank-consistency}
    \small
    \setlength{\tabcolsep}{3.2pt}
    \renewcommand{\arraystretch}{1.05}
    \begin{tabular}{@{}c@{\hspace{5pt}}c@{\hspace{5pt}}c@{\hspace{5pt}}cccc@{}}
        \toprule
        Model & Proj. & CR & Beacon & Raw & Spearman & MAE / Max \\
        \midrule
        \multirow{6}{*}{\textsc{Llama}}
        & K & 2 & 64.00 & 64.00 & 1.000 & 0.00 / 0 \\
        & K & 4 & 32.00 & 32.00 & 0.996 & 0.06 / 1 \\
        & K & 8 & 16.00 & 16.00 & 0.996 & 0.06 / 1 \\
        & V & 2 & 64.00 & 64.00 & 0.990 & 0.50 / 1 \\
        & V & 4 & 32.00 & 32.00 & 0.954 & 0.50 / 2 \\
        & V & 8 & 16.00 & 16.00 & 0.928 & 0.38 / 2 \\
        \midrule
        \multirow{6}{*}{\textsc{Qwen}}
        & K & 2 & 64.00 & 64.00 & 0.997 & 0.08 / 1 \\
        & K & 4 & 32.00 & 32.00 & 0.999 & 0.04 / 1 \\
        & K & 8 & 16.00 & 16.00 & 0.997 & 0.04 / 1 \\
        & V & 2 & 64.00 & 64.00 & 0.997 & 0.08 / 1 \\
        & V & 4 & 32.00 & 32.00 & 1.000 & 0.00 / 0 \\
        & V & 8 & 16.00 & 16.00 & 0.990 & 0.04 / 1 \\
        \bottomrule
    \end{tabular}
\end{table}

The results show strong agreement between Beacon and raw projections across models, projections, and compression ratios. In particular, the rank profiles achieve consistently high Spearman correlations, while the absolute discrepancies remain within one or two ranks in almost all cases. This supports using raw K/V projection spectra as a reliable proxy for layer-adaptive rank allocation before Beacon training.

\subsection{Interpretation of Attention Similarity Metrics}
\label{app:attention_similarity_metrics}

Figure~\ref{fig:attn_heatmap} visualizes the K-RoPE ablations, and Table~\ref{tab:heatmap_similarity} quantifies their similarity to the K-Kept setting. Let $A$ and $B$ be two averaged attention maps. All metrics are computed on the valid causal region
\begin{equation}
\Omega=\{(i,j):j\le i\}.
\end{equation}

\paragraph{Attention-map Pearson.}
We flatten $A$ and $B$ within $\Omega$ and compute
\begin{equation}
\rho_{\mathrm{attn}}
=
\mathrm{corr}
\left(
\{A_{ij}\}_{(i,j)\in\Omega},
\{B_{ij}\}_{(i,j)\in\Omega}
\right).
\end{equation}
Higher values indicate more similar global attention patterns.

\paragraph{Coarse-block Pearson.}
We average valid entries within coarse blocks to obtain $\widetilde{A}$ and $\widetilde{B}$, and compute
\begin{equation}
\rho_{\mathrm{block}}
=
\mathrm{corr}
\left(
\mathrm{vec}(\widetilde{A}),
\mathrm{vec}(\widetilde{B})
\right).
\end{equation}
Higher values indicate more similar macroscopic structures.

\paragraph{Distribution Divergence.}
For the two profile-based metrics below, we use Jensen-Shannon divergence:
\begin{align}
\mathrm{JS}(p,q)
&=
\frac{1}{2}\mathrm{KL}(p\|m)
+
\frac{1}{2}\mathrm{KL}(q\|m), \\
m
&=
\frac{1}{2}(p+q),
\end{align}
where $\mathrm{KL}(p\|m)=\sum_x p(x)\log\frac{p(x)}{m(x)}$. Lower JS values indicate more similar distributions.

\paragraph{Lag-profile JS Divergence.}
We aggregate attention mass by relative distance $\ell=i-j$ and normalize it into lag profiles $p_A(\ell)$ and $p_B(\ell)$:
\begin{equation}
D_{\mathrm{lag}}
=
\mathrm{JS}(p_A,p_B).
\end{equation}
Lower values indicate more similar near-range and long-range attention allocation.

\paragraph{Column-profile JS Divergence.}
We aggregate attention mass by key position $j$ and normalize it into column profiles $q_A(j)$ and $q_B(j)$:
\begin{equation}
D_{\mathrm{col}}
=
\mathrm{JS}(q_A,q_B).
\end{equation}
Lower values indicate more similar attended-key distributions.

\subsection{Derivation of Decoding Cost}
\label{app:cost_derivation}

We derive the dominant FLOPs of one decoding step. Let $H_q$ and $H_{kv}$ denote the numbers of query and KV heads, respectively, and let $g=H_q/H_{kv}$ be the GQA group size. Let $d$ be the per-head dimension, $D=H_{kv}d$ the concatenated KV dimension, $k$ the feature-wise compression ratio, and $N_b$ the number of cached beacon tokens. We consider a multi-head-coupled low-rank setting, where the latent KV dimension is $D/k$ over the concatenated KV space.

\paragraph{Reconstruction-based baseline.}
The reconstruction baseline stores the compressed beacon KV states $C_K,C_V\in\mathbb{R}^{N_b\times D/k}$ in the cache, and then expands them with the up-projections $U_K,U_V\in\mathbb{R}^{D/k\times D}$ before attention:
\begin{equation}
    K=C_KU_K,\qquad V=C_VU_V .
\end{equation}
Reconstructing both Keys and Values costs
\begin{equation}
    2\cdot N_b\frac{D}{k}D=\frac{2N_bD^2}{k}.
\end{equation}
After reconstruction, standard GQA attention is performed in the full feature space. The score computation and Value aggregation each cost $H_qN_bd=gN_bD$. In addition, applying K-side RoPE to reconstructed beacon Keys costs $N_bD$. Therefore,
\begin{equation}
    F_{\mathrm{recon}}\approx N_b\frac{2D^2}{k}+(1+2g)N_bD .
    \label{eq:app_recon_cost}
\end{equation}

\paragraph{SlimKV.}
SlimKV removes explicit K-side RoPE from beacon Keys and avoids full KV reconstruction. For each query head, the query is first projected into the latent space:
\begin{equation}
    \tilde q_h=q_hU_{K,h}^{\top}\in\mathbb{R}^{D/k}.
\end{equation}
This costs $dD/k$ per query head, or $H_qdD/k=gD^2/k$ over all query heads. The value-side expansion from the latent space back to the query-head output space has the same cost, giving a local projection cost of $2gD^2/k$.

Attention is then computed directly in the latent space. For each query head, both the score computation $\tilde q_hC_K^\top$ and the weighted aggregation over $C_V$ cost $N_bD/k$. Thus the latent attention cost is
\begin{equation}
    2H_qN_b\frac{D}{k}=\frac{2H_q}{k}N_bD .
\end{equation}
Combining the local projection and latent attention terms yields
\begin{equation}
    F_{\mathrm{SlimKV}}\approx \frac{2gD^2}{k}+\frac{2H_q}{k}N_bD .
    \label{eq:app_slimkv_cost}
\end{equation}

The key difference is that the reconstruction-based baseline incurs an $N_b$-scaled full-KV reconstruction cost, whereas SlimKV applies low-rank transformations only to the current-step query/output and keeps the dominant attention computation in the compressed latent space.

\paragraph{Typical-value analysis.}
We further instantiate Equations~\eqref{eq:app_recon_cost} and~\eqref{eq:app_slimkv_cost} with a representative GQA configuration, $H_q=32$, $H_{kv}=8$, $g=4$, $k=4$, and $D=1024$. For the reconstruction-based baseline, this gives
\begin{equation}
    F_{\mathrm{recon}}\approx \frac{1}{2}N_bD^2+9N_bD \approx 521N_bD ,
\end{equation}
where one factor of $D$ in the reconstruction term is instantiated as $1024$. For SlimKV, we have
\begin{equation}
    F_{\mathrm{SlimKV}}\approx 2D^2+16N_bD .
\end{equation}
The first term only comes from current-step query/output transformations and does not scale with $N_b$. It is dominated by the latent attention term when $2D^2\ll16N_bD$, i.e., $N_b\gg D/8$. For $D=1024$, this requires $N_b\gg128$, which is typically satisfied in long-context decoding. Therefore, we approximate
\begin{equation}
    F_{\mathrm{SlimKV}}\approx 16N_bD .
\end{equation}
Under this typical setting, the dominant arithmetic reduction is
\begin{equation}
    \frac{F_{\mathrm{recon}}}{F_{\mathrm{SlimKV}}}\approx \frac{521N_bD}{16N_bD}\approx 32.6\times .
\end{equation}
We emphasize that this comparison only reflects theoretical arithmetic cost: lower-order operations, kernel launch overheads, memory layout effects, and cache/memory-bandwidth limitations are omitted. Therefore, the arithmetic reduction should be interpreted as explaining the computational scaling trend rather than directly predicting measured wall-clock decoding speedup.

\subsubsection{Hybrid Attention with Raw-Token Tail}
\label{hybrid_attention}

In practical decoding, the cache may contain both compressed beacon tokens and a short tail of uncompressed raw tokens, e.g., newly generated tokens after the last compression step. We therefore use a hybrid attention computation: beacon tokens are processed in the latent KV space, while raw tokens are processed in the original feature space.

For a query head $h$, let $Q_h\in\mathbb{R}^{n\times d}$ be the current queries, $K_{r,h},V_{r,h}\in\mathbb{R}^{N_r\times d}$ be the raw-token KV cache, and $C_K,C_V\in\mathbb{R}^{N_b\times D/k}$ be the latent beacon KV cache. Let $U_{K,h},U_{V,h}\in\mathbb{R}^{D/k\times d}$ denote the corresponding low-rank up-projection matrices. We compute beacon and raw logits as
\begin{equation}
    S_{b,h}=\frac{(Q_hU_{K,h}^{\top})C_K^{\top}}{\sqrt{d}},\qquad S_{r,h}=\frac{Q_hK_{r,h}^{\top}}{\sqrt{d}}.
\end{equation}
The two logits are concatenated before softmax:
\begin{equation}
    [A_{b,h},A_{r,h}]=\mathrm{Softmax}\left([S_{b,h},S_{r,h}]\right),
\end{equation}
so beacon and raw tokens are normalized under the same attention distribution. The output is then
\begin{equation}
    O_h=(A_{b,h}C_V)U_{V,h}+A_{r,h}V_{r,h}.
    \label{eq:hybrid_attn}
\end{equation}

This is mathematically equivalent to reconstructing beacon KV states as $K_{b,h}=C_KU_{K,h}$ and $V_{b,h}=C_VU_{V,h}$, concatenating them with raw-token KV states, and applying standard attention:
\begin{equation}
    O_h=\mathrm{Softmax}\left(\frac{Q_h[K_{b,h};K_{r,h}]^{\top}}{\sqrt{d}}\right)[V_{b,h};V_{r,h}].
\end{equation}
The equivalence follows directly from associativity, i.e., $Q_h(C_KU_{K,h})^{\top}=(Q_hU_{K,h}^{\top})C_K^{\top}$ and $A_{b,h}(C_VU_{V,h})=(A_{b,h}C_V)U_{V,h}$. Thus, hybrid attention preserves the exact standard attention result while avoiding explicit full-dimensional reconstruction for beacon tokens.

\subsection{Implementation Details}
\label{app:impl_details}

\paragraph{Training cost and hardware.}
SlimKV freezes the backbone and updates only the beacon-branch QKV projections. All training runs are conducted with BF16 precision on 4 NVIDIA A100 80GB GPUs. For Llama-3.1-8B-Instruct, pre-training takes approximately 8 hours and fine-tuning takes approximately 6 hours, with about 50GB peak GPU memory usage. For Qwen2.5-14B-Instruct, pre-training takes approximately 14 hours and fine-tuning takes approximately 11 hours, with about 78GB peak GPU memory usage.

\paragraph{PALU settings.}
PALU jointly compresses KV cache by reducing feature dimension and numerical precision, so its total compression depends on both factors. We follow the original PALU settings whenever the target compression ratio is available. Otherwise, we choose the closest original setting and decrease feature compression only when necessary to meet the target ratio. This gives PALU a conservative configuration, since we observe that PALU is more sensitive to feature-dimensional compression than bit-width reduction. For 4$\times$, 8$\times$, 16$\times$, and 32$\times$ total compression, we use 1.5$\times$/6-bit, 1.5$\times$/3-bit, 2$\times$/3-bit, and 4$\times$/2-bit feature/precision settings, respectively.

\subsection{Detailed Efficiency Results}
\label{app:efficiency_details}

\paragraph{Decoding latency.}
Table~\ref{tab:appendix_latency_s} reports absolute decoding latency over 128 generated tokens. Recon denotes the reconstruction-based variant that retains beacon K-RoPE and materializes full-dimensional K/V states before attention. SlimKV consistently reduces both attention and end-to-end latency, confirming the benefit of reconstruction-free latent-space attention.

\begin{table}[h]
    \centering
    \scriptsize
    \setlength{\tabcolsep}{2.4pt}
    \caption{Detailed decoding latency on Llama-3.1-8B-Instruct over 128 generated tokens. Attn. and E2E denote cumulative attention and end-to-end decoding latency. All values are in seconds.}
    \resizebox{\columnwidth}{!}{%
    \begin{tabular}{cccccccccc}
        \toprule
        \multirow{2}{*}{\textbf{Comp.}} 
        & \multirow{2}{*}{\textbf{Method}}
        & \multicolumn{2}{c}{\textbf{32K}}
        & \multicolumn{2}{c}{\textbf{64K}}
        & \multicolumn{2}{c}{\textbf{96K}}
        & \multicolumn{2}{c}{\textbf{128K}} \\
        \cmidrule(lr){3-4}
        \cmidrule(lr){5-6}
        \cmidrule(lr){7-8}
        \cmidrule(lr){9-10}
        & & \textbf{Attn.} & \textbf{E2E}
          & \textbf{Attn.} & \textbf{E2E}
          & \textbf{Attn.} & \textbf{E2E}
          & \textbf{Attn.} & \textbf{E2E} \\
        \midrule
        1$\times$ & Full KV
        & 11.84 & 13.28
        & 23.97 & 25.40
        & 34.93 & 36.32
        & 46.64 & 48.08 \\
        \midrule
        \multirow{2}{*}{16$\times$}
        & Recon
        & 6.74 & 11.75
        & 8.52 & 15.46
        & 9.63 & 17.02
        & 8.73 & 16.79 \\
        & SlimKV
        & 6.11 & 10.93
        & 6.29 & 12.47
        & 6.87 & 13.79
        & 6.35 & 14.24 \\
        \midrule
        \multirow{2}{*}{32$\times$}
        & Recon
        & 7.80 & 11.56
        & 10.18 & 15.23
        & 9.11 & 13.95
        & 9.48 & 15.69 \\
        & SlimKV
        & 7.31 & 10.81
        & 9.43 & 14.51
        & 8.01 & 12.90
        & 7.65 & 13.78 \\
        \bottomrule
    \end{tabular}
    }
    \label{tab:appendix_latency_s}
\end{table}

\paragraph{TTFT and peak memory.}
Table~\ref{tab:appendix_ttft_memory} reports TTFT and PyTorch peak allocated memory. The uncompressed model runs out of memory at longer contexts, so the main text reports its longest feasible length of 11.5K tokens. Here we provide detailed measurements for SlimKV and the reconstruction-based variant at longer context lengths.

\begin{table}[h]
    \centering
    \scriptsize
    \setlength{\tabcolsep}{3.0pt}
    \caption{TTFT and peak allocated memory on Llama-3.1-8B-Instruct. TTFT is reported in seconds and memory in GB.}
    \resizebox{0.75\columnwidth}{!}{%
    \begin{tabular}{cccccc}
        \toprule
        \textbf{Comp.} & \textbf{Method} & \textbf{32K} & \textbf{64K} & \textbf{96K} & \textbf{128K} \\
        \midrule
        \multicolumn{6}{c}{\textbf{TTFT (s)}} \\
        \midrule
        \multirow{2}{*}{16$\times$}
        & Recon  & 16.59 & 46.38 & 89.15 & 153.74 \\
        & SlimKV & 16.16 & 44.80 & 87.12 & 152.57 \\
        \midrule
        \multirow{2}{*}{32$\times$}
        & Recon  & 11.28 & 28.34 & 50.13 & 82.73 \\
        & SlimKV & 10.92 & 27.77 & 48.62 & 79.96 \\
        \midrule
        \multicolumn{6}{c}{\textbf{Peak Memory (GB)}} \\
        \midrule
        \multirow{2}{*}{16$\times$}
        & Recon  & 25.73 & 29.77 & 33.80 & 37.84 \\
        & SlimKV & 25.72 & 29.76 & 33.80 & 37.84 \\
        \midrule
        \multirow{2}{*}{32$\times$}
        & Recon  & 23.42 & 25.26 & 27.09 & 28.92 \\
        & SlimKV & 23.41 & 25.25 & 27.08 & 28.91 \\
        \bottomrule
    \end{tabular}
    }
    \label{tab:appendix_ttft_memory}
\end{table}

Table~\ref{tab:appendix_ttft_memory} shows that higher compression ratios lead to clearly lower peak memory usage. SlimKV and the reconstruction-based variant have nearly identical peak memory because reconstruction changes only the computation flow, not the stored KV-cache size. SlimKV also achieves consistent but modest TTFT improvements over the reconstruction-based variant, mainly by avoiding key-side RoPE computation and full-Key materialization for beacon states.

\subsection{Task-wise Analysis on LongBench}
\label{app:taskwise_longbench}

We further analyze SlimKV across LongBench task categories. As shown in Table~\ref{tab:longbench_main}, SlimKV-f4 brings the most consistent gains on document QA tasks, including S-Doc and M-Doc, suggesting that RoPE-free beacon memory preserves key semantic evidence for retrieval-style reasoning while reducing token-wise compression pressure under the same total budget.

On summarization tasks, SlimKV gives smaller but stable gains, consistent with their stronger dependence on global semantic aggregation than exact token positions. Few-shot and code tasks show more backbone-dependent trends: SlimKV improves clearly on Qwen2.5-14B, while gains on Llama-3.1-8B are smaller; code tasks are also less uniform, likely due to their stronger reliance on local order, formatting, and structural cues.

Overall, RoPE-free beacon memory is especially effective for semantic aggregation and retrieval-oriented tasks, while fine-grained order-sensitive tasks remain more sensitive to the compression design.

\subsection{Needle-in-a-Haystack Evaluation Details}
\label{app:niah_heatmap}

Following the GPT-as-judge evaluation protocol in Activation Beacon~\citep{zhang-2025-ActivationBeacon}, we use GPT-5.4-mini~\citep{openai2026gpt54mini} as the judge with the following scoring prompt:

\begin{quote}
    \small
    Score 1: The answer is completely unrelated to the reference. \\
    Score 3: The answer has minor relevance but does not align with the reference. \\
    Score 5: The answer has moderate relevance but contains inaccuracies. \\
    Score 7: The answer aligns with the reference but has minor omissions. \\
    Score 10: The answer is completely accurate and aligns perfectly with the reference. \\
    Only respond with a numerical score.
\end{quote}

\begin{figure*}[p]
    \centering
    \includegraphics[width=\textwidth]{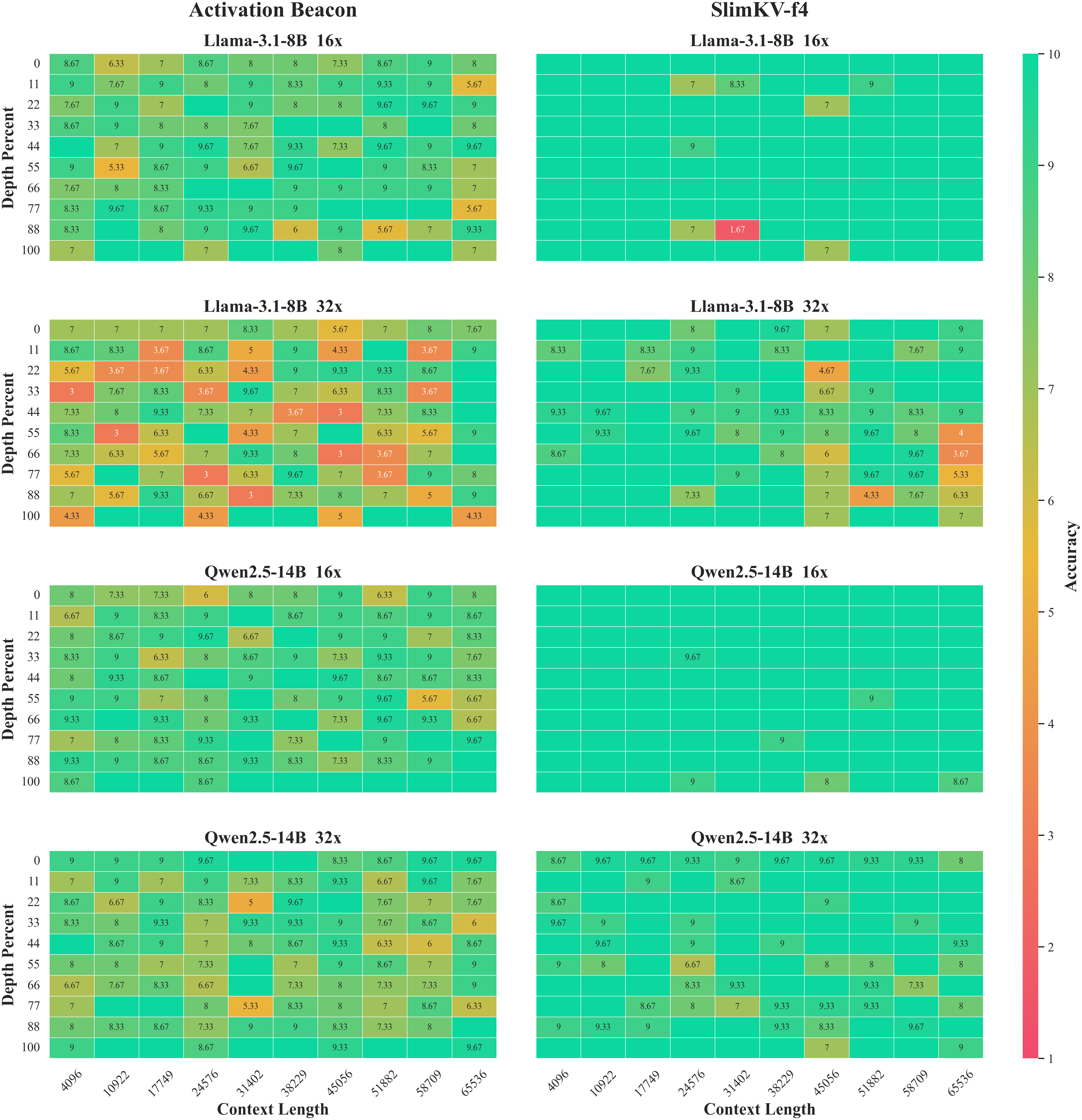}
    \caption{
    Complete Needle-in-a-Haystack heatmaps for Activation Beacon and SlimKV-f4, where f4 denotes 4$\times$ feature-wise compression. Results are shown on Llama-3.1-8B-Instruct and Qwen2.5-14B-Instruct under 16$\times$ and 32$\times$ total compression. Each cell reports retrieval accuracy at a given context length and needle depth.
    }
    \label{fig:niah_heatmap_appendix}
\end{figure*}

Figure~\ref{fig:niah_heatmap_appendix} provides the complete Needle-in-a-Haystack heatmaps across context lengths and needle depths, averaged over seeds 42, 43, and 44. SlimKV-f4 denotes 4$\times$ feature-wise compression. These results complement the main-text line plots and show that SlimKV-f4 maintains more stable retrieval performance than Activation Beacon.

\subsection{Comparison with MLA}

Both SlimKV and MLA~\citep{liu2024deepseek} use low-rank KV representations to avoid full-dimensional reconstruction. MLA compresses each token into a content latent $c_j^{KV}$ with a separate RoPE branch:
\begin{equation}
s_{ij}^{C}=(q_i^C U_K)(c_j^{KV})^\top ,
\end{equation}
\begin{equation}
s_{ij}^{\mathrm{MLA}}
=s_{ij}^{C}
+\langle \mathrm{RoPE}(q_i^R),\mathrm{RoPE}(k_j^R)\rangle .
\end{equation}
Thus, MLA preserves token granularity and an additional positional Key.

SlimKV instead jointly compresses tokens into beacon states and their K/V features:
\begin{equation}
s_{ij}^{\mathrm{SlimKV}}
=(\mathrm{RoPE}(q_i)U_K)C_{K,j}^{\top},
\end{equation}
\begin{equation}
o_i=(a_iC_V)U_V^{\top}.
\end{equation}
Beacon positional cues are acquired during RoPE-aware prefill, avoiding a separate positional-Key cache. Moreover, MLA is tightly coupled with the model architecture and requires costly base-model pretraining, while SlimKV retrofits pretrained GQA-RoPE backbones by training only beacon projections.

\subsection{Compatibility with ALiBi}

Beyond RoPE, we theoretically analyze SlimKV's compatibility with ALiBi~\citep{press2022train}. Since ALiBi adds positional bias after the QK product, for $K_b=C_KU_K^\top$,
\begin{equation}
S=QK_b^\top+B,
\end{equation}
\begin{equation}
S=(QU_K)C_K^\top+B.
\end{equation}
Thus, ALiBi is algebraically compatible with reconstruction-free latent-space attention without full-Key reconstruction.

\subsection{Licenses and Intended Use of Artifacts}
All backbones, training corpora, and evaluation benchmarks used in this work are publicly available, and we use them in accordance with their respective licenses, terms of use, and intended research purposes. Specifically, Llama-3.1-8B-Instruct is used under the Llama 3.1 Community License; Qwen2.5-14B-Instruct and Qwen3-30B-A3B-Instruct-2507 models, RedPajama under Apache 2.0; LongAlpaca under CC-BY-NC 4.0; BookSum under BSD-3-Clause; LongBench and Needle-in-a-Haystack under MIT. GPT-5.4-mini is accessed via the OpenAI API as an LLM-as-judge for NIAH evaluation, under the standard OpenAI Terms of Use.

\clearpage

\end{document}